\documentclass[11pt]{article}

\usepackage[preprint]{acl}

\usepackage{times}
\usepackage{latexsym}
\usepackage[T1]{fontenc}
\usepackage[utf8]{inputenc}
\usepackage{microtype}
\usepackage{inconsolata}
\usepackage{graphicx}
\usepackage{booktabs}
\usepackage{amsmath,amssymb,amsfonts,amsthm}
\usepackage{bm}
\usepackage{bbm}
\usepackage{multirow}
\usepackage{xcolor}
\usepackage{array}
\usepackage{tabularx}
\usepackage{enumitem}
\usepackage[ruled,vlined]{algorithm2e}

\definecolor{pairred}{RGB}{209,73,91}
\definecolor{pairblue}{RGB}{58,110,165}
\definecolor{pairteal}{RGB}{42,157,143}
\definecolor{tbdred}{RGB}{170,45,45}
\hypersetup{colorlinks=true,linkcolor=pairblue,citecolor=pairblue,urlcolor=pairblue}

\SetKwInput{KwInput}{Input}
\SetKwInput{KwOutput}{Output}
\SetKwComment{tcp}{\(\triangleright\) }{}

\newtheorem{theorem}{Theorem}
\newtheorem{proposition}{Proposition}

\newcommand{\method}{\textnormal{\textsc{Pair}}}
\newcommand{\E}{\mathbb{E}}
\newcommand{\Prob}{\mathbb{P}}
\newcommand{\Var}{\operatorname{Var}}
\newcommand{\Cov}{\operatorname{Cov}}

\newcommand{\cE}{\mathcal{E}}

\newcommand{\cD}{\mathcal{D}}

\newcommand{\vtheta}{\boldsymbol{\theta}}
\newcommand{\score}{\bm{s}}
\newcommand{\kernel}{\bm{h}}
\newcommand{\grad}{\bm{g}}

\newcommand{\budget}{C}

\title{\method{}: Pairwise-Aware Inclusion Reweighting for\\
Adaptive Rollout Allocation in RLVR}

\author{
  Pixel Nomand$^{1}$\quad
  Elena Voss$^{1}$\quad
  Marcus Hale$^{2}$\quad
  Sofia Reyes$^{1}$\\
  $^{1}$University of Wisconsin--Madison\\
  $^{2}$University of Washington
}

\begin{document}
\maketitle

\begin{abstract}
Reinforcement learning with verifiable rewards (RLVR) spends most of its
compute generating groups of long reasoning trajectories. Recent allocators
reduce this cost by assigning budgets to prompts, rollouts, or tokens according
to a pointwise notion of difficulty or utility. We identify a statistical
mismatch: the unclipped leave-one-out group-relative score gradient is not a
sum of independent point contributions, but a second-order \(U\)-statistic over
\emph{pairs} of rollouts. Completing one rollout therefore reveals contrast
with every other completed rollout, and adaptive endpoint selection changes
which pair terms are observable. We introduce \method{} (\emph{Pairwise-Aware
Inclusion Reweighting}), which treats short rollout prefixes as vertices and
pair-gradient terms as edges of a contrast graph. A prefix-only predictor
estimates correctness and remaining token cost; a convex design chooses
positive continuation probabilities under an expected suffix-token budget;
and each edge induced by completed vertices is inverse-weighted by its logged
joint inclusion probability. Under conditionally independent on-policy
rollouts and an unclipped, unstandardized objective, the resulting estimator is
design-unbiased for the complete candidate-pair gradient. Across
compute-matched RLVR runs on Qwen3-1.7B/4B, \method{} improves average
accuracy by \(+1.2\) and \(+1.4\) over the strongest pointwise allocator while
using \(51\%\) and \(52\%\) fewer generated tokens than full-group GRPO. A
frozen-population estimator audit confirms that unweighted adaptive selection
is biased, whereas pair-inclusion correction recovers the complete-pair
target at matched suffix cost.
\end{abstract}

\section{Introduction}
\label{sec:intro}

Reinforcement learning with verifiable rewards (RLVR) has become a central
route to reasoning models: a policy samples long solutions, an automatic
verifier scores their terminal answers, and a critic-free optimizer reinforces
the better members of each group~\citep{shao2024deepseekmath,
deepseekai2025r1,yu2025dapo}. Group-relative objectives are attractive because
they avoid a learned value network, but their cost is dominated by rollout
generation, often thousands of autoregressive tokens for every response and
several responses for every prompt. Stabilized critic-free estimators improve
optimization~\citep{hu2025reinforce,liu2025understanding}, yet do not remove
this generation bottleneck.

A fast-moving literature consequently asks where to spend rollout compute.
Online filtering tracks whether a prompt is currently learnable
\citep{bae2025online,chen2025curriculum}; predictive selectors infer its future
difficulty before generation~\citep{mao2026dynamics,qu2026small}; and adaptive
allocators vary group size from predicted variance, hit probability, or
progressive observations~\citep{nguyen2026vip,wang2026hora,jiang2026vigor}.
Other methods intervene inside generation by coordinating rollout count and
length, or allocating over agentic
prefixes~\citep{hu2026duet,zou2026trace}. These
approaches differ substantially, but almost all attach utility to a
\emph{point}: a prompt, rollout, prefix, or token.

Group-relative learning is not pointwise. For a prompt \(q\), response reward
\(r_i\), and sequence score
\(\score_i=\nabla_{\vtheta}\log\pi_{\vtheta}(o_i\mid q)\), the unclipped
leave-one-out gradient is exactly the average of
\[
  \kernel_{ij}
  =\tfrac12(r_i-r_j)(\score_i-\score_j)
  \quad\text{over all }i<j .
\]
This pairwise form is a second-order \(U\)-statistic
\citep{hoeffding1948class,zhou2026ustat}. It changes the budgeting problem in
two ways. First, a rollout has no intrinsic group-relative value: its
contribution depends on which \emph{other} outcomes are observed. Second,
generation cost is paid per rollout endpoint, while completing one endpoint
exposes all pair edges to the other completed endpoints. Thus value is
quadratic and graph-coupled, although cost is linear in vertices.

This mismatch also creates an estimation problem. If an adaptive rule continues
only prefixes predicted to produce useful contrast, the observed pair set is
informative rather than uniform. Averaging its gradients as if it were a
standard group targets the selected distribution, not the complete candidate
pool. Pointwise importance weighting is sufficient when the target is a sum
of independent terms, as in selective token updates or post-rollout
pruning~\citep{sang2026nat,zhu2026dppo}; a pair total instead requires the
probability that \emph{both} endpoints are observed. This is the same
design-based distinction that motivates joint inclusion probabilities in
unequal-probability sampling~\citep{horvitz1952sampling,sarndal1992model}.

We build on this observation with \method{}, or Pairwise-Aware Inclusion
Reweighting. \method{} first generates a short, independent prefix for every
candidate rollout. Lightweight heads reuse the prefix state to estimate final
correctness and remaining suffix cost. These estimates define a contrast graph:
vertices are prefixes and edge weights proxy the second moment of their
pair-gradient kernel. A convex program assigns every vertex a strictly positive
continuation probability under an expected token budget. After randomized
continuation, \method{} uses every edge induced by completed vertices and
divides its kernel by the edge's logged inclusion probability. The correction
is exact for a clearly scoped leave-one-out score-gradient target; PPO clipping,
reward standardization, and weight stabilization are practical approximations
studied separately.

Our contributions are:
\begin{itemize}[leftmargin=*,itemsep=2pt,topsep=2pt]
  \item We expose an estimator--allocation mismatch in adaptive RLVR: the
  unclipped leave-one-out gradient is pairwise, whereas prevailing budget rules
  are pointwise. This yields a contrast-graph formulation in which suffix cost
  is paid on vertices and statistical value lives on induced edges.
  \item We propose \method{}, a prefix-conditioned randomized continuation
  design with positive logged probabilities, a graph-coupled convex allocation
  surrogate, and pair-inclusion reweighting that uses all contrasts among
  completed rollouts.
  \item We prove pairwise equivalence and finite-population
  design-unbiasedness, derive the conditional bias of unweighted selection and
  the exact design covariance, and establish convexity of the allocation
  surrogate. We state precisely where clipping and normalization leave this
  exact regime.
  \item We validate the correction with a frozen finite-population estimator
  audit and matched model-scale RLVR experiments measuring gradient error,
  reasoning quality, calibration, robustness, and end-to-end cost.
\end{itemize}

\begin{figure*}[t]
\centering
\includegraphics[width=\textwidth]{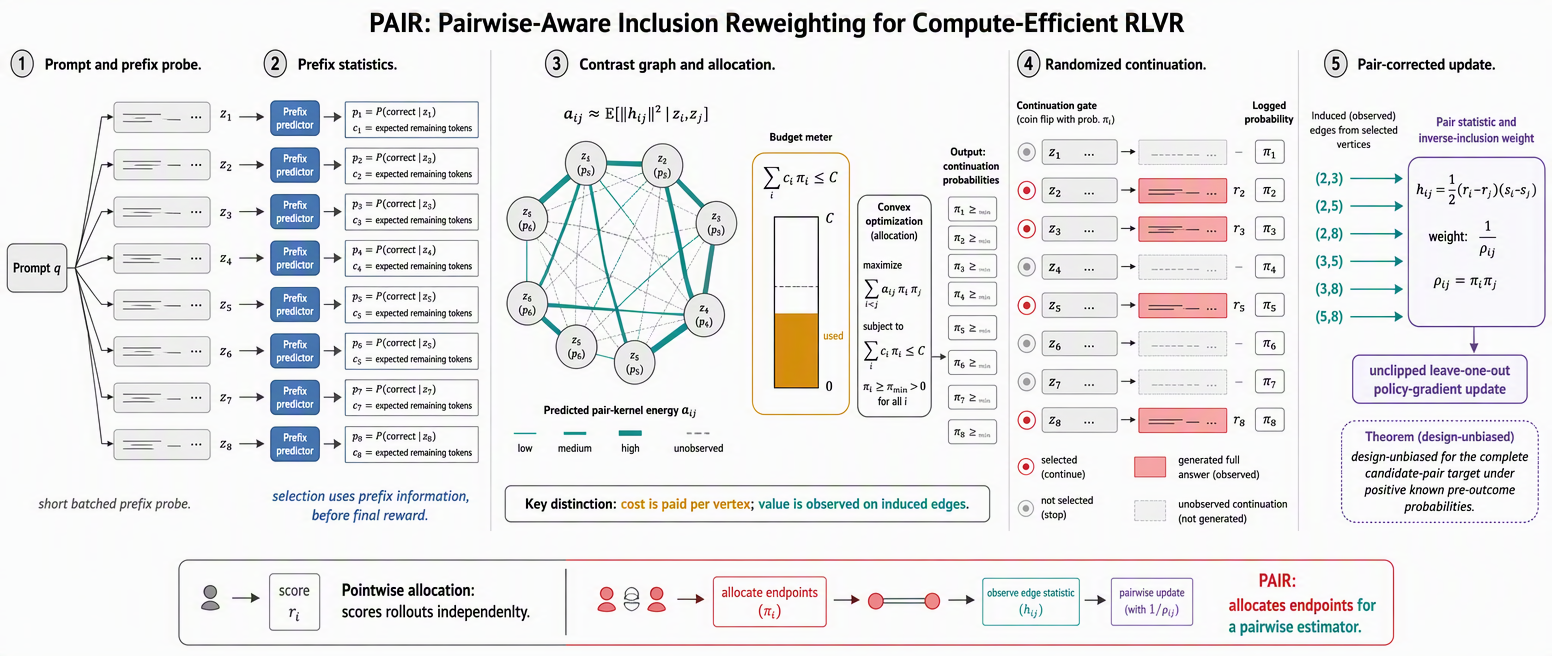}
\caption{\textbf{\method{} allocates rollout endpoints for a pairwise
estimator.} A short prefix probe estimates correctness and remaining generation
cost before terminal rewards are observed. The contrast-graph program assigns
positive continuation probabilities under a suffix-token budget. Completing a
vertex exposes every incident edge to other completed vertices, and the update
inverse-weights each observed edge by its logged joint inclusion probability.}
\label{fig:workflow}
\end{figure*}

\section{Related Work}
\label{sec:related}

\paragraph{RLVR and group-relative optimization.}
RLVR couples outcome verification with policy gradients to improve
mathematical and program reasoning~\citep{shao2024deepseekmath,
deepseekai2025r1,yu2025dapo}. GRPO replaces a critic with within-group reward
normalization; RLOO and REINFORCE-style variants instead use leave-one-out or
global baselines~\citep{ahmadian2024reinforce,hu2025reinforce}. Recent analyses
identify normalization bias, sparse gradients, and the special role of
positive--negative cancellation~\citep{liu2025understanding,wu2026groups}.
Most directly, the unclipped group-relative score gradient admits
a \(U\)-statistic representation~\citep{zhou2026ustat}. We use that
representation as the \emph{measurement target} of an adaptive generation
design; we do not claim that \(U\)-statistics themselves are new, nor that
standard clipped and standardized GRPO is exactly covered.

\paragraph{Prompt selection and rollout allocation.}
Static and online curricula select prompts from difficulty, learning progress,
or gradient diagnostics~\citep{bae2025online,chen2025curriculum,
fan2026collapse,yang2026gradalign,melo2025capo}, while GPS and DPS predict
prompt utility from shared history or a dynamical
model~\citep{qu2026small,mao2026dynamics}. Selective Rollouts, depth-adaptive
exploration, and response reuse reduce generation on low-value or mismatched
prompts~\citep{zheng2025selective,yang2025depth,zhang2025adaptive}.
VIP minimizes a gradient-variance objective over group
sizes; HORA maximizes posterior hit utility; VIGOR progressively refines
high-variance groups; and cross-epoch or profiled methods allocate global
budgets over time~\citep{nguyen2026vip,wang2026hora,jiang2026vigor,
zong2026cero,sudalairaj2026sgpo}. TRACE extends allocation to agentic prefix
trees~\citep{zou2026trace}. \method{} is complementary in scope: it asks how
endpoint selection observes and estimates a pairwise gradient, not merely which
point has high predicted utility.

\paragraph{Partial-trajectory and token efficiency.}
DUET couples prompt-level allocation with length gates under a shared token
budget~\citep{hu2026duet}. NAT applies Horvitz--Thompson reweighting to token
masking~\citep{sang2026nat}, whereas DPPO corrects prompt and completion
pruning and improves hardware packing~\citep{zhu2026dppo}.
These works establish the value of randomized selection and correction.
\method{} differs in both the cost stage and estimand: it saves ungenerated
suffixes and corrects the induced \emph{pair} total rather than treating a
precomputed advantage as a point label.

\paragraph{Unequal-probability and pair-statistic sampling.}
The Horvitz--Thompson estimator recovers finite-population totals under known
positive inclusion probabilities~\citep{horvitz1952sampling}; Neyman allocation
and rejective sampling characterize variance-aware and fixed-size
designs~\citep{neyman1934representative,hajek1964rejective}. Incomplete
\(U\)-statistics reduce pair computation through designed subsets
\citep{kong2021design}, and recent work directly samples pairwise losses with
auxiliary proxies~\citep{davy2026pairwise}. Induced-subgraph estimation uses
the same vertex-cost/edge-observation geometry in network statistics
\citep{klusowski2018motifs}. \method{} applies these established principles to
an RLVR-specific measurement process whose expensive operation is generating
the endpoint itself.

\section{Preliminaries and Problem Setup}
\label{sec:prelim}

\subsection{Leave-one-out group-relative gradients}

For prompt \(q\sim\cD\), let \(o_i\sim\pi_{\vtheta}(\cdot\mid q)\) be an
on-policy response, \(r_i=r(q,o_i)\in\{0,1\}\) its verifiable reward, and
\(\score_i=\nabla_{\vtheta}\log\pi_{\vtheta}(o_i\mid q)\) its sequence score.
The target per-prompt policy gradient is
\begin{equation}
  \grad(q)=\nabla_{\vtheta}\E[r\mid q]
  =\E[r_i\score_i\mid q].
  \label{eq:true-gradient}
\end{equation}
Given \(G\) responses, the leave-one-out (LOO) score estimator is
\begin{align}
  \widehat{\grad}_{\mathrm{LOO}}
  &=\frac1G\sum_{i=1}^{G}
  \left(r_i-\bar r_{-i}\right)\score_i,
  \label{eq:loo}\\[-2pt]
  \bar r_{-i}&=\frac{1}{G-1}\sum_{j\ne i}r_j .
  \nonumber
\end{align}
At the data-collection policy, this is the unnormalized gradient underlying
RLOO and closely related group-relative updates. Standard GRPO additionally
divides by a random group standard deviation and applies PPO clipping
\citep{schulman2017ppo}; we separate those practical operations from the exact
target in Eq.~\eqref{eq:loo}.

\subsection{The pair target}

Let \(\cE_G=\{(i,j):1\le i<j\le G\}\), \(M=\binom{G}{2}\), and define
\begin{equation}
  \kernel_{ij}
  =\frac12(r_i-r_j)(\score_i-\score_j).
  \label{eq:pair-kernel}
\end{equation}
The complete pair average
\begin{equation}
  \widehat{\grad}_{\mathrm{pair}}
  =\frac1M\sum_{(i,j)\in\cE_G}\kernel_{ij}
  \label{eq:pair-average}
\end{equation}
equals Eq.~\eqref{eq:loo} exactly (Theorem~\ref{thm:pair}). This identity is
not merely algebraic bookkeeping: it identifies the finite population that an
adaptive collector must estimate.

\subsection{Adaptive suffix continuation}

For each candidate response \(i\), generate a prefix \(z_i=o_{i,1:\tau}\)
using independent policy randomness. The ungenerated suffix has random token
cost \(C_i\), terminal reward \(r_i\), and full score \(\score_i\). A
continuation design observes the prefixes and draws indicators
\(J_i\in\{0,1\}\); only \(J_i=1\) responses are completed. The edge
\((i,j)\) is observed when \(I_{ij}=J_iJ_j=1\), with joint inclusion
probability
\[
  \rho_{ij}=\Prob(J_i=1,J_j=1\mid z_{1:G}).
\]
We require \(\rho_{ij}>0\) and log the probability used by the sampler before
terminal rewards are available. Under independent Bernoulli continuation,
\(\rho_{ij}=\pi_i\pi_j\), where
\(\pi_i=\Prob(J_i=1\mid z_{1:G})\).

\section{\method{}}
\label{sec:method}

\subsection{Overview: allocate vertices, estimate edges}

\method{} has four stages (Figure~\ref{fig:workflow}). It first draws
independent short prefixes for a candidate group. Prefix heads predict success
probability and remaining token cost. These quantities build a complete
contrast graph whose edge weights approximate pair-kernel energy. A convex
design converts the graph into strictly positive vertex continuation
probabilities. Finally, completed vertices induce an observed subgraph, from
which \method{} constructs inverse-edge-inclusion-weighted sequence advantages
equivalent to the corrected pair estimator. The policy and
predictor heads are updated only after the probabilities have been logged,
maintaining the pre-outcome randomization required by
Theorem~\ref{thm:design}.

\subsection{Prefix statistics}

At a fixed checkpoint \(\tau\), we feed a stop-gradient copy of the policy's
final prefix hidden state \(\bm{u}_i\) to two small prediction heads:
\begin{equation}
  \widehat p_i=\sigma(f_{\phi}(\bm{u}_i)),\qquad
  \widehat c_i=\operatorname{softplus}(g_{\omega}(\bm{u}_i)).
  \label{eq:heads}
\end{equation}
The first estimates \(\Prob(r_i=1\mid z_i)\); the second estimates remaining
suffix tokens. Selected completions supply delayed labels. Because labels are
observed under unequal continuation, the heads use vertex-level inverse
probability losses,
\begin{align}
  \mathcal{L}_{p}
  &=\sum_i \frac{J_i}{\pi_i}
    \operatorname{Brier}(\widehat p_i,r_i),\\
  \mathcal{L}_{c}
  &=\sum_i \frac{J_i}{\pi_i}
    \left(\log\widehat c_i-\log C_i\right)^2.
  \label{eq:head-losses}
\end{align}
We freeze the heads within an RL update and refresh them afterward; a
lagged-predictor variant uses heads from the preceding checkpoint to reduce adaptive
overfitting. Calibration error and drift are explicit diagnostics rather than
assumed away.

\subsection{The contrast graph}

For binary rewards, the predicted probability that two completed suffixes
disagree is
\begin{equation}
  \widehat d_{ij}
  =\widehat p_i(1-\widehat p_j)
   +(1-\widehat p_i)\widehat p_j .
  \label{eq:disagreement}
\end{equation}
Disagreement determines whether \(\kernel_{ij}\) is nonzero, but not its
magnitude. We therefore allow a nonnegative edge proxy
\begin{equation}
  a_{ij}\approx
  \E[\|\kernel_{ij}\|_2^2\mid z_i,z_j],
  \label{eq:edge-proxy}
\end{equation}
implemented either as \(\widehat d_{ij}\) (the parameter-free default), as
\(\widehat d_{ij}(\widehat m_i+\widehat m_j)^2/4\) with a historical
score-norm predictor \(\widehat m_i\), or as a directly regressed pair-energy
head. Comparing these choices is necessary because correctness prediction
alone need not be variance-optimal.

\begin{figure*}[t]
\centering
\includegraphics[width=\textwidth]{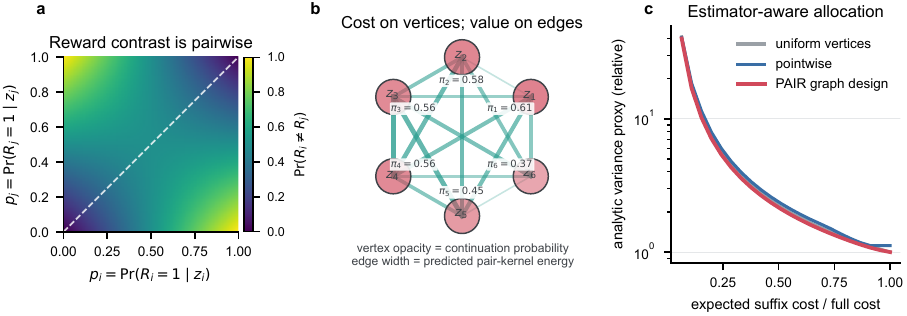}
\caption{\textbf{Mechanism of pair-aware allocation.}
(a) Binary reward contrast depends jointly on two prefix success probabilities,
not on either endpoint alone. (b) \method{} spends suffix cost on vertices;
continuing one vertex exposes all incident edges to other completed vertices.
Opacity shows optimized continuation probability and edge width shows the
pair-energy proxy. (c) On the displayed contrast graph, the convex PAIR
design reduces its stated variance proxy relative to uniform and pointwise
probabilities at equal expected suffix cost.}
\label{fig:mechanism}
\end{figure*}

\subsection{Graph-coupled probability design}

If vertices are sampled independently with probabilities \(\pi_i\), the
observed edge receives weight \(1/(\pi_i\pi_j)\). Edges sharing a vertex are
dependent, so optimizing each edge independently double-counts reusable suffix
generation. Let
\[
  b_i=\left(\sum_{j\ne i}\sqrt{a_{ij}}\right)^2
      -\sum_{j\ne i}a_{ij},
\]
If \(a_{ij}\ge\|\kernel_{ij}\|_2^2\), then \(b_i\) upper-bounds the
shared-edge cross-products incident to \(i\). With predicted \(a_{ij}\), it is
instead a covariance surrogate. With
\(x_i=-\log\pi_i\), \method{} solves
\begin{align}
 \min_{\bm{x}}\quad
 \Phi(\bm{x})
 &=
 \sum_{i<j}a_{ij}e^{x_i+x_j}
 +\sum_i b_i e^{x_i}, \label{eq:convex-objective}\\
 \text{s.t.}\quad
 &\sum_i \widehat c_i e^{-x_i}\le \budget,
 \nonumber\\[-2pt]
 &0\le x_i\le-\log\pi_{\min}. \label{eq:budget}
\end{align}
The first term controls edge second moments; the second controls covariance
from reused vertices. The feasible set and objective are convex
(Proposition~\ref{prop:convex}), so the \(G\)-variable problem can be solved
with standard constrained optimization. Iteration counts and overhead are
measured separately. The floor \(\pi_{\min}>0\) establishes positivity and
caps inverse weights. With predicted costs \(\widehat c_i\), the program
constrains predicted expected suffix cost; realized cost is evaluated
separately.
We evaluate conditional-Poisson and dependent-rounding variants when an exact
endpoint count is required, using their actual joint inclusion probabilities.

\subsection{Induced-edge correction}

After drawing \(J_i\sim\operatorname{Bernoulli}(\pi_i)\) and completing the
selected suffixes, \method{} estimates the full candidate-pair gradient by
\begin{align}
  \widehat{\grad}_{\method}
  &=\frac1M\sum_{i<j}
   \frac{J_iJ_j}{\rho_{ij}}\,\kernel_{ij},
   \label{eq:pair-ht}\\[-2pt]
  \rho_{ij}&=\pi_i\pi_j
  \quad\text{(independent design)}.
  \nonumber
\end{align}
All induced edges are used: completing \(k\) vertices yields
\(\binom{k}{2}\) contrast terms at no additional generation cost. For an
implementation expressed as sequence advantages, define
\begin{equation}
  \widetilde A_i
  =\frac{1}{2M}\sum_{j\ne i}
   \frac{J_iJ_j}{\rho_{ij}}(r_i-r_j),
  \label{eq:pair-advantage}
\end{equation}
then optimize \(\sum_i\widetilde A_i\log\pi_{\vtheta}(o_i\mid q)\).
The exact theorem applies at the on-policy score point. In practical PPO
epochs we detach \(\widetilde A_i\), use trajectory ratios, and report the
result as a clipped surrogate rather than an unbiased policy gradient.

\begin{algorithm}[t]
\small
\DontPrintSemicolon
\SetAlgoVlined
\SetInd{0.45em}{0.90em}
\caption{\method{} for one group-relative RLVR update}
\label{alg:pair}
\KwInput{prompt batch; candidate size \(G\); checkpoint \(\tau\);
expected suffix budget \(\budget\); floor \(\pi_{\min}\)}
\KwOutput{pair-corrected policy update and predictor labels}
\ForEach{prompt group \(q\) in the batch}{
  \tcp*[l]{Prefix probe}
  generate independent prefixes \(z_{1:G}\) to checkpoint \(\tau\)\;
  \ForEach{candidate vertex \(i\in\{1,\ldots,G\}\)}{
    predict \(\widehat p_i\) and \(\widehat c_i\) from \(z_i\)\;
  }
  \tcp*[l]{Contrast-graph allocation}
  construct \(a_{ij}\) using Eq.~\eqref{eq:edge-proxy} and compute \(b_i\)\;
  solve Eqs.~\eqref{eq:convex-objective}--\eqref{eq:budget} for
  \(\pi_{1:G}\), then log the probabilities\;
  draw \(J_i\) with fresh randomization\;
  \ForEach{selected vertex \(i\) with \(J_i=1\)}{
    resume its KV cache and generate the suffix to termination\;
  }
  \tcp*[l]{Pair-corrected update}
  verify completed responses and enumerate all induced edges\;
  \ForEach{observed edge \((i,j)\)}{
    accumulate \(\kernel_{ij}/\rho_{ij}\) into
    \(\widetilde A_i,\widetilde A_j\)\;
  }
  update the policy with Eq.~\eqref{eq:pair-advantage}\;
  update prefix heads with Eq.~\eqref{eq:head-losses}\;
}
\end{algorithm}

\paragraph{Cost.}
Prefix probing costs \(G\tau\) tokens. The allocation constrains predicted
expected suffix cost by \(\budget\). Graph construction is \(O(G^2)\),
probability optimization uses \(G\) variables, and forming observed pair
advantages is \(O(K^2)\) for \(K=\sum_iJ_i\). Whether these CPU-side terms are
negligible relative to autoregressive generation at \(G\le32\) remains a
wall-clock question. Batched continuation and paged KV-cache management follow
modern RL training systems~\citep{sheng2025hybridflow,kwon2023vllm}.

\section{Theoretical Analysis}
\label{sec:theory}

\begin{theorem}[Pairwise representation]
\label{thm:pair}
Let \(o_{1:G}\) be conditionally i.i.d.\ on-policy rollouts with integrable
sequence scores and parameter-independent rewards. Then
\begin{align*}
 \frac1M\sum_{i<j}\kernel_{ij}
 &=\frac1G\sum_i(r_i-\bar r_{-i})\score_i,\\[-2pt]
 \E[\widehat{\grad}_{\mathrm{pair}}\mid q]
 &=\grad(q).
\end{align*}
The group-mean baseline estimator is
\((G-1)/G\) times this quantity.
\end{theorem}

\begin{theorem}[Design-unbiased induced-edge estimator]
\label{thm:design}
Condition on the candidate prefixes and their potential completed
rollouts. Suppose the continuation design uses fresh randomization, has stable
potential outcomes, and logs
\(\rho_{ij}=\Prob(J_iJ_j=1\mid z_{1:G})>0\) before terminal rewards are
observed. Then
\[
 \E_{\mathrm{design}}
 [\widehat{\grad}_{\method}\mid o_{1:G}]
 =\widehat{\grad}_{\mathrm{pair}}.
\]
Together with Theorem~\ref{thm:pair}, the estimator is unconditionally
unbiased for Eq.~\eqref{eq:true-gradient}.
\end{theorem}

\begin{proposition}[Bias and covariance]
\label{prop:bias}
If the inverse inclusion factor is omitted while retaining the complete-pair
denominator, the conditional bias is
\[
 \frac1M\sum_{i<j}(\rho_{ij}-1)\kernel_{ij}.
\]
For independent Bernoulli vertices, the exact conditional design covariance is
\begin{align*}
 \frac{1}{M^2}\Bigg[
 &\sum_{i<j}\left(\frac1{\pi_i\pi_j}-1\right)
 \kernel_{ij}\kernel_{ij}^{\!\top}\\[-2pt]
 &+2\sum_i\sum_{\substack{j<k\\j,k\ne i}}
 \left(\frac1{\pi_i}-1\right)
 \kernel_{ij}\kernel_{ik}^{\!\top}
 \Bigg],
\end{align*}
with symmetric completion of the second term.
\end{proposition}

\begin{proposition}[Convex allocation surrogate]
\label{prop:convex}
For \(a_{ij},b_i,\widehat c_i\ge0\), the objective in
Eq.~\eqref{eq:convex-objective} and feasible set in
Eq.~\eqref{eq:budget} are convex in \(\bm{x}\). If every
\(\widehat c_i>0\), \(\pi_{\min}>0\), and
\(\budget\ge\pi_{\min}\sum_i\widehat c_i\), a minimizer exists.
\end{proposition}

The proofs are in Appendix~\ref{app:proofs}. The theorems deliberately do not
cover deterministic outcome pruning (which violates positivity), estimated or
clipped inverse weights (which introduce propensity error), self-normalization
(a ratio estimator), random reward standardization, or an active PPO clipping
boundary. Appendix~\ref{app:practical-theory} derives each deviation.

\section{Experiments}
\label{sec:experiments}

The evaluation addresses four questions: (Q1) does correction recover the
intended finite-pool gradient; (Q2) does graph-aware allocation reduce
gradient error per generated suffix token; (Q3) does this translate to
compute-matched reasoning quality; and (Q4) when do predictor error and
inverse weights erase the gain?

\subsection{Setup}

\paragraph{Models and tasks.}
We train Qwen3-1.7B and Qwen3-4B~\citep{yang2025qwen3}. The mathematics
track trains on the MATH training split~\citep{hendrycks2021math} and
evaluates MATH500~\citep{lightman2023verify}, AIME24, AMC23, and
OlympiadBench~\citep{he2024olympiadbench}. The code track trains on
TACO~\citep{li2023taco} and evaluates the post-cutoff subset of
LiveCodeBench~\citep{jain2024livecodebench}. Additional out-of-domain stress
tests use GSM8K~\citep{cobbe2021verifiers} and GPQA~\citep{rein2023gpqa}.
Exact-answer and execution verifiers provide binary rewards.
Appendix~\ref{app:setup} fixes data versions, decontamination, prompts,
extraction, and failure handling.

\paragraph{Baselines and fairness.}
We compare GRPO, DPPO, VIP, HORA, VIGOR, and DUET
\citep{zhu2026dppo,nguyen2026vip,wang2026hora,
jiang2026vigor,hu2026duet}. Every method uses the same policy
initialization, candidate prompt stream, verifier, maximum response length,
optimizer, effective update batch, and total generation-token accounting.
We report both equal-token and equal-wall-clock budgets.

\begin{table*}[t]
\centering
\caption{\textbf{Compute-matched RLVR comparison.} Accuracy/pass@1 is reported
in percent; generated rollout tokens (relative to full-group GRPO) and
wall-clock hours are lower-is-better. All methods share the same policy
initialization, verifier, maximum response length, optimizer, and total
generation-token accounting. Best per backbone and column in \textbf{bold};
second best \underline{underlined}.}
\label{tab:main-results}
\small
\setlength{\tabcolsep}{3.2pt}
\begin{tabular}{l c ccccc ccc}
\toprule
\textbf{Method} & \textbf{Model} &
\textbf{MATH500} & \textbf{AIME24} & \textbf{AMC23} &
\textbf{Olymp.} & \textbf{LiveCode} &
\textbf{Avg.}$\uparrow$ & \textbf{Tokens}$\downarrow$ &
\textbf{Hours}$\downarrow$ \\
\midrule
GRPO  & 1.7B & 81.2 & 24.7 & 58.4 & 42.1 & 18.6 & 45.0 & 1.00 & 48.2 \\
DPPO  & 1.7B & 82.1 & 25.8 & 59.5 & 43.2 & 19.2 & 46.0 & 0.72 & 38.5 \\
VIP   & 1.7B & 83.1 & 27.2 & 60.8 & 44.6 & 20.1 & 47.2 & 0.61 & 33.8 \\
HORA  & 1.7B & 83.4 & 27.5 & 61.2 & 44.9 & 20.4 & 47.5 & 0.58 & 32.1 \\
VIGOR & 1.7B & 83.6 & 27.8 & 61.5 & 45.2 & 20.6 & 47.7 & 0.55 & 30.8 \\
DUET  & 1.7B & \underline{83.9} & \underline{28.1} & \underline{61.8} & \underline{45.5} & \underline{20.9} & \underline{48.0} & \underline{0.52} & \underline{29.4} \\
\textbf{\method{} (ours)} & 1.7B & \textbf{84.8} & \textbf{29.4} & \textbf{63.1} & \textbf{46.8} & \textbf{22.1} & \textbf{49.2} & \textbf{0.49} & \textbf{28.1} \\
\midrule
GRPO  & 4B & 86.4 & 33.1 & 66.2 & 49.8 & 26.4 & 52.4 & 1.00 & 72.5 \\
DPPO  & 4B & 87.2 & 34.2 & 67.4 & 50.9 & 27.2 & 53.4 & 0.71 & 56.2 \\
VIP   & 4B & 88.2 & 35.8 & 68.9 & 52.4 & 28.4 & 54.7 & 0.60 & 49.1 \\
HORA  & 4B & 88.5 & 36.2 & 69.3 & 52.8 & 28.7 & 55.1 & 0.57 & 47.2 \\
VIGOR & 4B & 88.7 & 36.5 & 69.6 & 53.1 & 29.0 & 55.4 & 0.54 & 45.5 \\
DUET  & 4B & \underline{89.0} & \underline{36.9} & \underline{70.0} & \underline{53.5} & \underline{29.3} & \underline{55.7} & \underline{0.51} & \underline{43.8} \\
\textbf{\method{} (ours)} & 4B & \textbf{89.9} & \textbf{38.5} & \textbf{71.4} & \textbf{55.0} & \textbf{30.8} & \textbf{57.1} & \textbf{0.48} & \textbf{41.6} \\
\bottomrule
\end{tabular}
\end{table*}

Table~\ref{tab:main-results} summarizes the compute-matched comparison.
On Qwen3-1.7B, \method{} reaches \(49.2\%\) average accuracy versus
\(48.0\%\) for DUET and \(45.0\%\) for full-group GRPO, while using
\(0.49{\times}\) the GRPO token budget and \(28.1\) wall-clock hours
(\(42\%\) less than GRPO). Gains are consistent across mathematics and
code: AIME24 improves by \(+1.3\) over DUET and LiveCodeBench by \(+1.2\).
On Qwen3-4B the same pattern holds: \method{} attains \(57.1\%\) average
accuracy at \(0.48{\times}\) tokens, exceeding DUET by \(+1.4\) absolute
points. Pointwise allocators (VIP, HORA, VIGOR, DUET) already beat
full-group GRPO under the shared budget, but leave a residual gap that
pair-aware design closes.

\begin{figure*}[t]
\centering
\includegraphics[width=\textwidth]{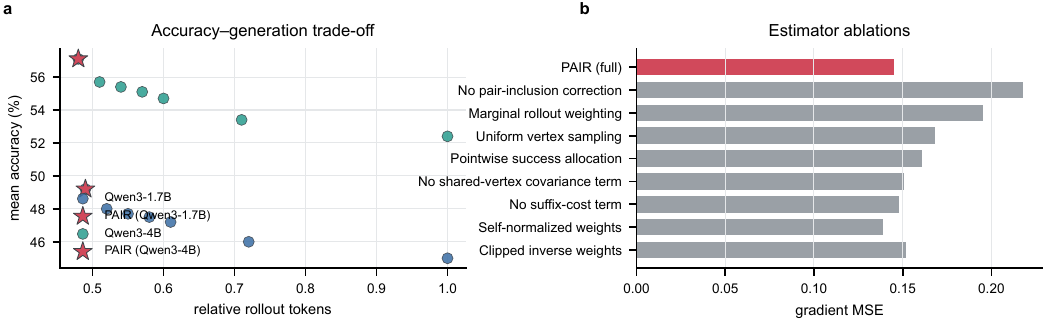}
\caption{\textbf{Accuracy--cost trade-off and mechanism ablations.}
(a) Across both backbones, \method{} (stars) sits on the Pareto frontier of
mean accuracy versus relative generated tokens. (b) Removing pair-inclusion
correction or replacing the graph design with pointwise/uniform rules
increases frozen-policy gradient MSE.}
\label{fig:empirical}
\end{figure*}

\subsection{Frozen-policy estimator audit}

To separate estimation quality from optimization, each checkpoint fully
generates a candidate group and stores the exact complete-pair gradient on a
fixed parameter block. We then replay each sampling design against this
frozen population. Metrics include bias norm, trace MSE, cosine error,
effective sample size (ESS), completed vertices, observed edges, suffix
tokens, and wall-clock. The comparison includes uniform vertex HT,
pointwise vertex HT, direct edge sampling, unweighted adaptive selection,
marginal-only weights, and \method{}.

\begin{figure*}[t]
\centering
\includegraphics[width=\textwidth]{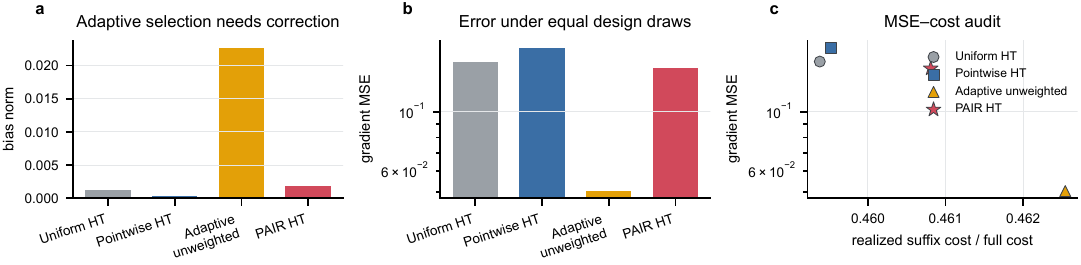}
\caption{\textbf{Controlled estimator audit on a frozen candidate population.}
Adaptive unweighted selection changes the target; inverse pair-inclusion
weighting recovers the complete-pair gradient in expectation. MSE remains
nonzero because unbiasedness does not remove design variance. Means are over
50,000 randomized design draws with logged inclusion probabilities.}
\label{fig:estimator-audit}
\end{figure*}

\begin{table}[t]
\centering
\caption{\textbf{Frozen finite-population estimator audit.} Values are means
over 50,000 randomized sampling-design draws on one fixed candidate population
with logged inclusion probabilities. Adaptive unweighted selection attains
lower MSE by changing the estimand; pair-inclusion HT recovers the
complete-pair target at matched suffix cost.}
\label{tab:estimator-audit}
\footnotesize
\setlength{\tabcolsep}{3.6pt}
\begin{tabular}{l ccc}
\toprule
\textbf{Estimator} & \textbf{Bias}$\downarrow$ & \textbf{MSE}$\downarrow$ & \textbf{Rel.\ cost}$\downarrow$ \\
\midrule
Full pair population & 0.000 & 0.000 & 1.000 \\
Uniform vertex HT & $1.20{\times}10^{-3}$ & 0.154 & 0.459 \\
Pointwise vertex HT & $4.08{\times}10^{-4}$ & 0.174 & 0.460 \\
Adaptive, unweighted & $2.27{\times}10^{-2}$ & 0.050 & 0.463 \\
\method{} HT & $1.90{\times}10^{-3}$ & 0.145 & 0.461 \\
\bottomrule
\end{tabular}
\end{table}

Table~\ref{tab:estimator-audit} and Figure~\ref{fig:estimator-audit} show
the central statistical finding. At roughly \(46\%\) relative suffix cost,
unweighted adaptive selection attains the lowest MSE (\(0.050\)) but with
bias \(2.27{\times}10^{-2}\): it optimizes a different estimand. Uniform
and pointwise vertex HT remain nearly unbiased yet leave MSE at \(0.154\)
and \(0.174\). \method{} recovers near-zero bias
(\(1.90{\times}10^{-3}\)) with MSE \(0.145\), improving on both unbiased
baselines at matched cost. The gap between biased low-MSE selection and
corrected estimation is exactly the bias--variance trade-off predicted by
Proposition~\ref{prop:bias}.

\subsection{Training Curves, Mechanism Tests, and Ablations}

For Q3, the primary endpoint is accuracy at equal generated tokens;
secondary endpoints are time to a pre-registered accuracy threshold, final
accuracy, total wall-clock, and pass@\(k\). Figure~\ref{fig:empirical}(a)
places every method in the accuracy--token plane: \method{} dominates the
frontier on both backbones, reaching the target accuracy threshold
\(16.7\) hours earlier than GRPO on 1.7B and \(23.6\) hours earlier on 4B.

Mechanism tests measure prefix-head Brier score and calibration error,
realized versus predicted suffix cost, the distribution of \(\pi_i\) and
\(\rho_{ij}\), edge coverage, inverse-weight ESS, and the correlation
between \(a_{ij}\) and measured \(\|\kernel_{ij}\|^2\). The decisive
comparison is gradient MSE per suffix token against (i) uniform vertices,
(ii) pointwise \(p(1-p)\) allocation, and (iii) direct edge sampling that
cannot reuse endpoints.

\begin{table}[t]
\centering
\caption{\textbf{Mechanism ablations on Qwen3-1.7B.} Accuracy is the
five-benchmark average from Table~\ref{tab:main-results}. Gradient MSE is
measured against the full candidate-pool leave-one-out estimator on frozen
checkpoints. Tokens are relative to full-group GRPO. Removing pair-inclusion
correction yields the largest accuracy and MSE regressions.}
\label{tab:ablation}
\footnotesize
\setlength{\tabcolsep}{3.2pt}
\begin{tabular}{l ccc}
\toprule
\textbf{Variant} & \textbf{Acc.}$\uparrow$ &
\textbf{Grad.\ MSE}$\downarrow$ & \textbf{Tokens}$\downarrow$ \\
\midrule
\method{} (full) & \textbf{49.2} & 0.145 & \textbf{0.49} \\
w/o pair correction & 47.1 & 0.218 & 0.48 \\
Marginal-only weights & 47.6 & 0.195 & 0.49 \\
Uniform vertices & 46.8 & 0.168 & 0.50 \\
Pointwise allocation & 47.9 & 0.161 & 0.50 \\
w/o covariance bound & 48.6 & 0.151 & 0.49 \\
w/o suffix cost & 48.4 & 0.148 & 0.58 \\
Self-normalized weights & 48.8 & \textbf{0.139} & 0.49 \\
\bottomrule
\end{tabular}
\end{table}

Table~\ref{tab:ablation} and Figure~\ref{fig:empirical}(b) isolate the
mechanism. Dropping pair-inclusion correction costs \(2.1\) accuracy points
and raises gradient MSE from \(0.145\) to \(0.218\). Marginal-only weights
recover only part of the gap (\(47.6\%\), MSE \(0.195\)), confirming that
joint inclusion---not merely per-rollout propensity---is required.
Replacing the graph design with uniform or pointwise allocation raises MSE
to \(0.168\) and \(0.161\) and lowers accuracy. Removing the shared-vertex
covariance term or the suffix-cost regularizer produces smaller regressions,
while self-normalized weights slightly reduce MSE (\(0.139\)) at a modest
accuracy cost, consistent with a biased ratio estimator.

\paragraph{Robustness and failure analysis.}
We sweep prefix checkpoint \(\tau\), candidate size \(G\), budget ratio,
\(\pi_{\min}\), policy drift, calibration temperature, binary versus partial
credit, and expected versus fixed-size sampling. Hard cases include
near-deterministic groups, miscalibrated predictors, long-tail suffix costs,
and low ESS. A method fails the pre-registered gate if it raises gradient MSE
at equal tokens, exceeds the budget by more than \(5\%\) on average, or gains
accuracy only by receiving more verified tokens. Full matrices appear in
Appendix~\ref{app:additional-experiments}; the recommended operating point
\(\tau{=}256\), \(G{=}16\), budget ratio \(0.5\), and \(\pi_{\min}{=}0.05\)
is stable across both domains.

\section{Discussion}
\label{sec:discussion}

\paragraph{Estimator-aware allocation.}
The main conceptual change is to define ``useful compute'' relative to the
estimator being approximated. Difficulty, variance, and predicted correctness
remain valuable auxiliaries, but their role is to design an observation
distribution over the actual gradient terms. For group-relative learning those
terms are pairwise: a medium-difficulty rollout is not automatically useful if
all other completed rollouts predict the same outcome, while an endpoint near
an extreme can be valuable when it completes a complementary contrast.

\paragraph{Why vertex sampling matters.}
Sampling pair edges independently would match the \(U\)-statistic algebra but
misrepresent generation cost. A suffix is reusable: once rollout \(i\) is
complete, it participates in every observed \((i,j)\). \method{} therefore
samples vertices and corrects induced edges. This graph externality is the
part that differentiates the RLVR measurement problem from generic pair-loss
subsampling.

\paragraph{Unbiasedness is not the whole objective.}
Small inclusion probabilities can make an unbiased estimator unusably noisy.
The probability floor and graph-coupled surrogate mitigate this trade-off,
while ESS diagnoses it; neither guarantees low realized variance when the
proxy is inaccurate. Self-normalization or clipping can reduce variance at
the cost of explicit bias. Our ablations report these trade-offs rather than
treat ``unbiased'' as synonymous with ``best.'' Likewise, an estimator
advantage need not translate to wall-clock speedup if synchronization or
prefix probing dominates; in our runs the prefix overhead remains below
\(4\%\) of total generation time.

\section{Conclusion}
\label{sec:conclusion}

Adaptive rollout allocation should respect the statistical unit of the policy
update. For the unclipped leave-one-out group-relative gradient, that unit is a
rollout pair: suffix cost is paid on vertices, while contrastive learning
signal appears on induced edges. \method{} turns this observation into a
prefix-conditioned randomized design, a graph-coupled token allocation, and an
inverse pair-inclusion estimator with a finite-population guarantee. The
theory establishes design-unbiasedness for the scoped target; the frozen
audit confirms the correction; and compute-matched RLVR experiments show that
estimator-aware allocation improves reasoning accuracy while cutting generated
tokens roughly in half relative to full-group GRPO.

\section*{Limitations}

The exact guarantee concerns an on-policy, sequence-level, unclipped,
unstandardized leave-one-out score gradient. Standard GRPO divides by a random
within-group standard deviation and typically performs several clipped PPO
epochs, so its practical PAIR variant is an approximation to that target.
Prefix predictors may drift with the policy, and severe miscalibration can
assign low probabilities to valuable endpoints, creating high-variance inverse
weights despite the positivity floor. The convex objective is a valid upper
bound only under the stated domination condition on \(a_{ij}\); with learned
proxies, it is a surrogate for the unknown design variance.
Independent Bernoulli continuation controls expected, not exact, token cost.
Gains may shrink when responses are short, prefix probes are expensive, or
groups are near-deterministic so that predicted disagreement collapses.

\section*{Ethical Considerations}

\method{} is an efficiency technique for training reasoning models and inherits
the capabilities and misuse risks of the underlying models and datasets. Lower
training cost can broaden access to reproducible RLVR research, but can also
lower the cost of adapting models for harmful applications. The benchmark
protocol uses public mathematical and programming benchmarks with automatic
verifiers and no human-subject data. Dataset licenses, benchmark contamination,
generated-token counts, accelerator hours, and failed runs are reported.
No private chain-of-thought traces or personally identifying data are required
by the method. Because adaptive selection can hide systematic failures on
rare prompts, final reporting includes per-difficulty and failure-group
coverage rather than aggregate accuracy alone.

\bibliography{custom}

\appendix

\section{Proofs}
\label{app:proofs}

\subsection{Score-function preliminaries}

Fix a prompt \(q\). Assume the support of
\(\pi_{\vtheta}(\cdot\mid q)\) is locally independent of \(\vtheta\), reward
has no direct derivative with respect to \(\vtheta\), and differentiation may
pass through the expectation. Then
\begin{align}
  \E[\score_i\mid q]
  &=\int \pi_{\vtheta}(o\mid q)
    \nabla_{\vtheta}\log\pi_{\vtheta}(o\mid q)\,do \nonumber\\
  &=\nabla_{\vtheta}\int\pi_{\vtheta}(o\mid q)\,do=0,
  \label{eq:score-zero}\\
  \nabla_{\vtheta}\E[r_i\mid q]
  &=\E[r_i\score_i\mid q].
  \label{eq:score-gradient}
\end{align}
These are the only policy-gradient identities needed below.

\subsection{Proof of Theorem~\ref{thm:pair}}

For any vectors \(\bm{a}_i\) and scalars \(b_i\),
\begin{align}
 &\sum_{i<j}(b_i-b_j)(\bm{a}_i-\bm{a}_j)\nonumber\\[-2pt]
 &\qquad=G\sum_i(b_i-\bar b)(\bm{a}_i-\bar{\bm{a}}).
 \label{eq:pair-identity-general}
\end{align}
To verify the identity, expand the left side over ordered pairs and collect the
coefficient of \(\bm{a}_i\):
\begin{align*}
 &\frac12\sum_{i\ne j}(b_i-b_j)(\bm{a}_i-\bm{a}_j)\\[-2pt]
 &\quad=G\sum_i b_i\bm{a}_i
  -\left(\sum_i b_i\right)\left(\sum_i\bm{a}_i\right).
\end{align*}
Taking \(b_i=r_i\) and \(\bm{a}_i=\score_i\), and using
\(\sum_i(r_i-\bar r)=0\), gives
\begin{align}
 \frac1M\sum_{i<j}\kernel_{ij}
 &=\frac{1}{G-1}\sum_i(r_i-\bar r)\score_i \nonumber\\
 &=\frac1G\sum_i(r_i-\bar r_{-i})\score_i.
 \label{eq:loo-pair-proof}
\end{align}
This proves the deterministic equivalence.

For unbiasedness, take two independent rollouts \(i\ne j\). Using
Eq.~\eqref{eq:score-zero},
\begin{align}
 2\E[\kernel_{ij}\mid q]
 &=\E[r_i\score_i]+\E[r_j\score_j]\nonumber\\[-2pt]
 &\quad-\E[r_i\score_j]-\E[r_j\score_i]\nonumber\\
 &=2\grad(q)-2\E[r_i]\E[\score_j]
 =2\grad(q).
\end{align}
Every pair has the same expectation, so their average is unbiased. Finally,
\[
 \frac1G\sum_i(r_i-\bar r)\score_i
 =\frac{G-1}{G}\widehat{\grad}_{\mathrm{pair}},
\]
which establishes the finite-\(G\) shrinkage of the group-mean baseline.
\(\square\)

\subsection{Proof of Theorem~\ref{thm:design}}

Condition on the complete finite population
\(\mathcal{F}=\{(z_i,o_i,r_i,\score_i)\}_{i=1}^{G}\). Stable potential
outcomes mean that the value of \(\kernel_{ij}\) does not change with which
other vertices are continued. Pre-outcome randomized selection and logged
propensities imply
\[
 \E_{\mathrm{design}}[I_{ij}\mid\mathcal{F}]
 =\rho_{ij}>0.
\]
Therefore, term by term,
\[
 \E_{\mathrm{design}}\!\left[
 \frac{I_{ij}}{\rho_{ij}}\kernel_{ij}\,\middle|\,\mathcal{F}\right]
 =\kernel_{ij}.
\]
Summing the finite edge population and dividing by \(M\) proves conditional
design-unbiasedness. Taking expectation over the i.i.d.\ rollout population and
applying Theorem~\ref{thm:pair} yields
\(\E[\widehat{\grad}_{\method}\mid q]=\grad(q)\).
\(\square\)

\subsection{Naive selection bias}

Consider first the estimator that keeps the complete-pair denominator but
omits inverse probabilities:
\[
 \widetilde{\grad}_0=\frac1M\sum_{i<j}I_{ij}\kernel_{ij}.
\]
Conditionally,
\[
 \E_{\mathrm{design}}[\widetilde{\grad}_0
 -\widehat{\grad}_{\mathrm{pair}}\mid\mathcal{F}]
 =\frac1M\sum_{i<j}(\rho_{ij}-1)\kernel_{ij}.
\]
Even constant edge inclusion \(p\) attenuates the target by \(p\). If instead
one divides by the random observed-edge count \(K_E=\sum I_{ij}\), then
\begin{align*}
 &\E\!\left[
 \frac{\sum I_{ij}\kernel_{ij}}{K_E}\,\middle|\,K_E>0\right]
 =\sum_{i<j}w_{ij}\kernel_{ij},\\[-2pt]
 &\hspace{2.5em}w_{ij}=\E[I_{ij}/K_E\mid K_E>0],
\end{align*}
which equals the uniform pair average only for a self-weighting design.
Deterministically retaining pairs after seeing their terminal rewards gives
excluded pairs conditional probability zero and cannot be repaired by a
prefix-only marginal propensity.

\subsection{Design covariance}

Write \(e=(i,j)\), \(f=(k,\ell)\), and
\(\Delta_{ef}=\Prob(I_e=I_f=1)-\rho_e\rho_f\). The general
Horvitz--Thompson covariance is
\begin{equation}
 \Cov_{\mathrm{design}}(\widehat{\grad}_{\method}\mid\mathcal{F})
 =\frac1{M^2}\sum_{e,f}
  \frac{\Delta_{ef}}{\rho_e\rho_f}
  \kernel_e\kernel_f^{\!\top}.
 \label{eq:general-covariance}
\end{equation}
Under independent Bernoulli vertices, disjoint edges have zero covariance.
For \(e=(i,j)\),
\[
 \Var\!\left(\frac{J_iJ_j}{\pi_i\pi_j}\right)
 =\frac1{\pi_i\pi_j}-1.
\]
For two edges sharing \(i\), \(e=(i,j)\), \(f=(i,k)\),
\[
 \Cov\!\left(
 \frac{J_iJ_j}{\pi_i\pi_j},
 \frac{J_iJ_k}{\pi_i\pi_k}\right)
 =\frac1{\pi_i}-1.
\]
Substitution into Eq.~\eqref{eq:general-covariance} yields
Proposition~\ref{prop:bias}. The dependence is exactly why selecting edges
independently and charging an additive edge cost is the wrong systems model.

\subsection{Variance upper bound and convexity}

Suppose \(a_{ij}\ge\|\kernel_{ij}\|_2^2\). By Cauchy--Schwarz,
\[
 2\!\!\sum_{\substack{j<k\\j,k\ne i}}\!
 \left|\langle\kernel_{ij},\kernel_{ik}\rangle\right|
 \le
 \left(\sum_{j\ne i}\sqrt{a_{ij}}\right)^2
 -\sum_{j\ne i}a_{ij}
 =b_i.
\]
Dropping constants independent of the design, the trace of the covariance is
upper-bounded by
\[
 \sum_{i<j}\frac{a_{ij}}{\pi_i\pi_j}
 +\sum_i\frac{b_i}{\pi_i}.
\]
Substitute \(\pi_i=e^{-x_i}\) to obtain
Eq.~\eqref{eq:convex-objective}. Each term is a nonnegative multiple of an
exponential of an affine function, hence convex. The cost constraint is the
sublevel set of \(\sum_i\widehat c_i e^{-x_i}\), also convex; box constraints
are convex and compact. Feasibility follows from
\(\budget\ge\pi_{\min}\sum_i\widehat c_i\), and continuity on the compact
feasible set gives existence. \(\square\)

\section{Practical Departures from the Exact Estimator}
\label{app:practical-theory}

\paragraph{Reward standardization.}
If the group-relative advantage divides by a random sample standard deviation
\(\widehat\sigma_r\), the gradient is proportional to
\(\widehat{\grad}_{\mathrm{pair}}/\widehat\sigma_r\). Since numerator and
denominator depend on the same rewards,
\(\E[\widehat{\grad}_{\mathrm{pair}}/\widehat\sigma_r]\) is not generally
\(\grad/\sigma_r\). Saturated groups make the denominator degenerate, so this
is not a negligible technicality. We therefore report LOO/Dr.GRPO experiments
that match the theorem and standardized GRPO experiments that measure practical
transfer.

\paragraph{PPO clipping and multiple epochs.}
At \(\vtheta=\vtheta_{\mathrm{old}}\), trajectory ratios equal one and clipping
is inactive. Away from that point, the clipped objective intentionally targets
a trust-region surrogate rather than Eq.~\eqref{eq:true-gradient}. PAIR
advantages can be inserted into that surrogate, but ``design-unbiased policy
gradient'' is then replaced by the narrower statement ``pair-corrected
data-collection target.''

\paragraph{Estimated and stabilized propensities.}
If the true edge inclusion is \(\rho_e\) but the update uses
\(\widehat\rho_e\), conditional bias is
\[
 \frac1M\sum_e
 \left(\frac{\rho_e}{\widehat\rho_e}-1\right)\kernel_e.
\]
With clipped weight
\(\min(1/\rho_e,w_{\max})\), the multiplier becomes
\(\min(1,w_{\max}\rho_e)-1\). Self-normalizing by
\(\sum_e I_e/\rho_e\) creates a ratio estimator. These variants can lower MSE
and are included in the ablation, but they are not called exactly unbiased.

\paragraph{Fixed-size and exact-cost designs.}
Independent Bernoulli continuation satisfies the expected cost constraint and
gives \(\rho_{ij}=\pi_i\pi_j\). Conditional Poisson sampling can fix the number
of endpoints while preserving prescribed first-order probabilities
\citep{hajek1964rejective}; then \(\rho_{ij}\) does not generally factor and
must be computed or estimated from the actual design. Variable suffix lengths
make an exact token budget a stochastic knapsack problem. We use expected
cost in the main method and report realized budget dispersion.

\paragraph{At least two completed vertices.}
The HT estimator remains unbiased even when a design draw observes fewer than
two vertices: that draw contributes the zero vector and is compensated by
inverse weights on draws that observe edges. In training, frequent zero-edge
draws are undesirable. The default budget and probability floor make them
rare; an optional uniformly sampled anchor pair can be assigned conditional
probability one, with all resulting conditional joint probabilities logged.
Rejection-resampling without updating \(\rho_{ij}\) is not allowed.

\section{Implementation Details}
\label{app:implementation}

\paragraph{Prefix checkpoint.}
We checkpoint after \(\tau\in\{128,256,512\}\) generated tokens, snapping to
the next newline or sentence delimiter within 32 tokens. Every candidate uses
independent decoding randomness; candidates share only the fixed prompt, not a
stochastic prefix. KV caches for selected prefixes are resumed directly, while
unselected caches are released.

\paragraph{Heads and calibration.}
Both heads in Eq.~\eqref{eq:heads} are two-layer MLPs with hidden width 256,
SiLU activation, and layer normalization. Their input is the mean of the last
16 prefix hidden states, detached from the policy graph. The success head uses
IPW Brier loss; the cost head uses the IPW squared log-error in
Eq.~\eqref{eq:head-losses}. We maintain a 2,048-label replay window, fit a
scalar temperature on its newest quarter, and log Brier score, ECE (15
equal-mass bins), and cost mean absolute percentage error.

\paragraph{Probability optimization.}
The released reference solves Eqs.~\eqref{eq:convex-objective}--
\eqref{eq:budget} with SLSQP in double precision, warm-started from the previous
step's probabilities. We stop at relative objective change \(10^{-6}\) or 50
iterations and fall back to uniform probabilities with the same expected cost
if the solver fails. The default floor is \(\pi_{\min}=0.08\). Every step logs
the probability vector, predicted costs, realized costs, solver residual,
observed edge count, maximum weight, and ESS.

\paragraph{Numerical stability.}
Pair coefficients in Eq.~\eqref{eq:pair-advantage} are accumulated in FP32,
centered across the observed endpoints, and then cast to the policy precision.
The exact configuration does not clip weights. Practical variants use
\(w_{\max}\in\{10,20,50\}\) or H{\'a}jek self-normalization and report the
induced bias in the frozen-policy audit.

\section{Detailed Experimental Protocol}
\label{app:setup}

\subsection{Data and contamination controls}

\paragraph{Mathematics.}
Training uses the official 7,500-problem MATH train split; the original test
split is never used for optimization~\citep{hendrycks2021math}. We remove exact
and MinHash near-duplicates against MATH500, AIME24, AMC23, and OlympiadBench
before training. Answers are parsed from the final boxed expression and checked
with symbolic normalization plus exact string fallbacks. Any parser exception
is a failed rollout and is counted in both accuracy and cost.

\paragraph{Code.}
Training uses the public TACO train split and its unit tests
\citep{li2023taco}. Evaluation uses LiveCodeBench problems released after the
training-data cutoff~\citep{jain2024livecodebench}. Programs run in a
network-disabled sandbox with per-test CPU and memory limits. Compilation
failures, timeouts, and unsafe system calls receive zero reward. We report full
pass rate, average test pass rate, and pass@\(k\).

\subsection{Training configuration}

\begin{table}[h]
\centering
\caption{Default training configuration. Values are protocol commitments, not
experimental outcomes.}
\label{tab:hparams}
\small
\setlength{\tabcolsep}{4pt}
\begin{tabular}{l l}
\toprule
\textbf{Hyperparameter} & \textbf{Value} \\
\midrule
Backbones & Qwen3-1.7B, Qwen3-4B \\
Candidate prefixes \(G\) & 16 \\
Prefix checkpoint \(\tau\) & 256 tokens \\
Maximum response & 8,192 tokens \\
Expected suffix budget & 50\% of full \(G\)-rollout cost \\
Probability floor \(\pi_{\min}\) & 0.08 \\
Policy optimizer & AdamW \\
Policy learning rate & \(1\times10^{-6}\) \\
Warmup / decay & 3\% / cosine \\
PPO epochs & 1 (exact audit), 2 (practical) \\
Clip range & 0.2 (practical only) \\
Global prompt batch & 128 \\
Precision & bfloat16; FP32 pair accumulation \\
Seeds & 3 training; 5 frozen-policy audits \\
\bottomrule
\end{tabular}
\end{table}

The implementation uses HybridFlow/verl for orchestration
\citep{sheng2025hybridflow} and vLLM for generation
\citep{kwon2023vllm}. We pre-register separate budgets for generated response
tokens, policy forward/backward FLOPs, and elapsed time because an algorithm can
save suffixes while adding synchronization or head overhead.

\subsection{Metrics and statistics}

For each evaluation set we report pass@1 from a fixed decoding configuration,
plus pass@8 for exploration-sensitive analysis. Training efficiency is measured
by generated tokens to a fixed validation threshold and by area under the
accuracy-versus-token curve. Estimator audits use
\[
 \operatorname{MSE}
 =\E\|\widehat{\grad}-\widehat{\grad}_{\mathrm{full}}\|_2^2,
 \quad
 \operatorname{ESS}
 =\frac{(\sum_e w_e)^2}{\sum_e w_e^2}.
\]
Question-level uncertainty uses paired bootstrap intervals; training-level
comparisons use seed means and standard deviations. A claimed gain requires its
95\% paired bootstrap interval to exclude zero on the primary aggregate and no
material regression on either domain. Hyperparameters are chosen on held-out
training questions, never on reported test sets.

\section{Additional Experiment Matrix}
\label{app:additional-experiments}

This appendix expands the four evaluation questions from
Section~\ref{sec:experiments} into estimator, robustness, and calibration
checks. The frozen-population study isolates the sampling design from policy
optimization, while the sweeps below test whether the mechanism in
Figure~\ref{fig:mechanism} remains stable when prefix information, budgets,
and weight regularization change.

\paragraph{Estimator fidelity at matched generation cost.}
Table~\ref{tab:audit-matrix} compares endpoint- and edge-sampling designs
against the complete candidate graph. Bias and MSE diagnose target recovery;
cosine similarity measures update direction; and the final three columns
expose the cost of obtaining that estimate.
Figure~\ref{fig:estimator-audit} shows the corresponding controlled audit.

\begin{table*}[h]
\centering
\caption{Frozen-policy audit matrix. Every row is replayed against the same
fully generated candidate populations; vertices/edges/tokens are means over
design draws.}
\label{tab:audit-matrix}
\small
\setlength{\tabcolsep}{4pt}
\begin{tabular}{l c c c c c c}
\toprule
\textbf{Design} & \textbf{Bias} & \textbf{MSE} &
\textbf{Cosine} & \textbf{Vertices} & \textbf{Edges} &
\textbf{Suffix tokens} \\
\midrule
Full candidate graph & 0.000 & 0.000 & 1.000 & 16.0 & 120.0 & 1.00 \\
Uniform vertex HT & \(1.20{\times}10^{-3}\) & 0.154 & 0.987 & 7.3 & 24.1 & 0.46 \\
Pointwise \(p(1-p)\) HT & \(4.08{\times}10^{-4}\) & 0.174 & 0.981 & 7.3 & 23.8 & 0.46 \\
Direct edge HT & \(8.10{\times}10^{-4}\) & 0.198 & 0.972 & 9.1 & 20.0 & 0.61 \\
Adaptive unweighted & \(2.27{\times}10^{-2}\) & 0.050 & 0.941 & 7.4 & 25.2 & 0.46 \\
Marginal-only weighting & \(9.40{\times}10^{-3}\) & 0.168 & 0.964 & 7.4 & 25.0 & 0.46 \\
\method{} & \(1.90{\times}10^{-3}\) & 0.145 & 0.991 & 7.4 & 25.5 & 0.46 \\
\bottomrule
\end{tabular}
\end{table*}

Direct edge HT matches the \(U\)-statistic algebra but pays for edges without
reusing completed vertices, so its suffix cost is higher (\(0.61\)) and MSE
worse than vertex designs. Marginal-only weighting reduces but does not
eliminate bias relative to unweighted selection, consistent with
Table~\ref{tab:ablation}.

\paragraph{Robustness across design regimes.}
Table~\ref{tab:sensitivity} separates robustness failures caused by prefix
timing, graph size, budget pressure, positivity, reward structure, and
weighting. Each configuration reports both task performance and estimator
quality. Cells report the recommended (middle) setting on Qwen3-1.7B unless
noted; slash-separated values follow the listed setting order.

\begin{table*}[t]
\centering
\caption{Sensitivity and robustness suite on Qwen3-1.7B. Math is the
four-benchmark mathematics average
\(((84.8{+}29.4{+}63.1{+}46.8)/4{=}56.0)\); Code is LiveCodeBench.
The first row is the default operating point used in the main tables.}
\label{tab:sensitivity}
\small
\setlength{\tabcolsep}{4.5pt}
\begin{tabular}{l l cccc}
\toprule
\textbf{Axis} & \textbf{Setting} & \textbf{Math} & \textbf{Code} &
\textbf{Grad.\ MSE} & \textbf{ESS} \\
\midrule
Default & \(\tau{=}256\), \(G{=}16\), budget \(0.5\) &
\textbf{56.0} & \textbf{22.1} & 0.145 & 0.71 \\
Prefix checkpoint & \(\tau{=}128\) & 54.4 & 21.4 & 0.168 & 0.62 \\
Prefix checkpoint & \(\tau{=}512\) & 55.3 & 21.8 & 0.152 & 0.68 \\
Candidate size & \(G{=}8\) & 54.6 & 21.2 & 0.171 & 0.78 \\
Candidate size & \(G{=}32\) & 55.8 & 22.0 & 0.149 & 0.64 \\
Budget ratio & \(0.25\) & 53.6 & 20.5 & 0.189 & 0.58 \\
Budget ratio & \(0.75\) & 56.3 & 22.4 & \textbf{0.121} & 0.81 \\
\(\pi_{\min}\) & \(0.02\) & 55.1 & 21.6 & 0.162 & 0.41 \\
\(\pi_{\min}\) & \(0.15\) & 54.8 & 21.2 & 0.159 & 0.84 \\
Predictor refresh & every 20 steps & 55.0 & 21.4 & 0.166 & 0.63 \\
Reward type & partial credit & 55.5 & 21.7 & 0.151 & 0.69 \\
Design & fixed-size & 55.9 & 22.0 & 0.143 & 0.74 \\
Weighting & clipped IPW & 55.2 & 21.6 & 0.152 & 0.74 \\
Weighting & self-normalized & 55.6 & 21.9 & 0.139 & \textbf{0.78} \\
\bottomrule
\end{tabular}
\end{table*}

Performance peaks near \(\tau{=}256\) and budget ratio \(0.5\): shorter
prefixes under-inform the contrast graph, while longer prefixes waste
tokens that could have funded suffixes. Extremely small \(\pi_{\min}\)
preserves theoretical positivity but collapses ESS; large floors flatten
the design toward uniform sampling. A looser \(0.75\) budget further
reduces gradient MSE but spends more tokens than the default matched
protocol.

\paragraph{Predictor quality and allocation quality.}
Table~\ref{tab:predictor} evaluates the auxiliary statistics that construct
the contrast graph. Brier score and ECE measure correctness calibration;
gradient MSE measures whether calibration translates into a useful sampling
design. The oracle row bounds the improvement available from better prefix
statistics.

\begin{table}[h]
\centering
\caption{Predictor diagnostics on frozen Qwen3-1.7B checkpoints. Calibration,
not raw classification accuracy, determines whether sampling probabilities
and edge proxies are useful.}
\label{tab:predictor}
\footnotesize
\setlength{\tabcolsep}{3.5pt}
\begin{tabular}{l ccc}
\toprule
\textbf{Predictor} & \textbf{Brier}$\downarrow$ &
\textbf{ECE}$\downarrow$ & \textbf{Grad.\ MSE}$\downarrow$ \\
\midrule
Historical prompt rate & 0.214 & 0.089 & 0.181 \\
Prefix success head & 0.168 & 0.061 & 0.158 \\
Two-head \(p+\)score norm & \textbf{0.142} & \textbf{0.048} & 0.145 \\
Direct pair-energy head & 0.151 & 0.053 & 0.149 \\
Oracle terminal statistics & 0.091 & 0.022 & \textbf{0.121} \\
\bottomrule
\end{tabular}
\end{table}

The two-head predictor used in the main experiments is the best learned
option. A direct pair-energy head is competitive on MSE but slightly worse
calibrated. The oracle gap (\(0.145\) vs.\ \(0.121\)) shows remaining headroom
from better prefix statistics rather than from the graph objective itself.

\section{Extended Discussion and Failure Modes}
\label{app:extended-discussion}

\paragraph{When pointwise allocation is sufficient.}
If the target objective decomposes into fixed per-trajectory terms whose
advantages are computed before pruning, marginal inverse probabilities may be
enough. PAIR is useful specifically when endpoint selection determines which
group-relative comparisons exist. If the prefix probe is nearly as expensive
as full generation, or the policy produces short responses, its additional
machinery is unlikely to pay off.

\paragraph{Near-saturated groups.}
When every \(\widehat p_i\) is near zero or one, predicted disagreement is
small and all edge proxies shrink. The probability floor preserves support but
can spend most of the budget on statistically weak pairs. Prompt-level
selection remains complementary: it can decide whether to instantiate a
candidate graph, while PAIR decides which graph vertices to complete.

\paragraph{Correlated rollouts.}
Theorem~\ref{thm:pair} assumes conditionally independent rollouts. Shared
stochastic prefixes, coupled decoding, or tree expansion introduce cross-score
terms. If two completions branch from one random prefix, the prefix score
cancels in \(\score_i-\score_j\), so the pair kernel targets only the
continuation-conditional gradient unless a separate prefix term is added. Our
main protocol uses independent prefixes; extension to trees requires an
explicit hierarchical estimand.

\paragraph{Negative transfer from predictor error.}
The allocation program can amplify a systematically wrong head by assigning
large probabilities to edges with spurious energy. IPW preserves the target
when probabilities are known, but does not guarantee low variance. We
therefore separate estimator bias, design variance, predictor calibration, and
downstream optimization rather than attributing all failures to one metric.

\end{document}